\documentclass{article}
\usepackage{arxiv}
\usepackage{graphicx}
\usepackage{amsmath}
\usepackage{amssymb}
\usepackage[utf8]{inputenc}
\usepackage{latexsym}
\usepackage{booktabs}
\usepackage{array,multirow}
\usepackage{mathtools}
\usepackage{colortbl,xcolor}
\usepackage{subfig}
\usepackage[labelfont=bf,font=small]{caption}
\usepackage[parfill]{parskip}
\usepackage{xifthen}
\usepackage[pagebackref=true,breaklinks=true,colorlinks,bookmarks=false]{hyperref}

\title{HaarNet: Large-scale Linear-Morphological Hybrid Network for RGB-D Semantic Segmentation}
\author{
 Rick Groenendijk \\
  University of Amsterdam \\
  Science Park 900, 1098XH Amsterdam \\
  \texttt{r.w.groenendijk@uva.nl} \\
   \And
 Leo Dorst \\
  University of Amsterdam \\
  Science Park 900, 1098XH Amsterdam \\
  \texttt{l.dorst@uva.nl} \\
  \And
 Theo Gevers \\
  University of Amsterdam \\
  Science Park 900, 1098XH Amsterdam \\
  \texttt{th.gevers@uva.nl} \\
}

\newcolumntype{a}{>{\columncolor{lightgray}}l}
\newcommand*{\belowrulesepcolor}[1]{
  \noalign{
    \kern-\belowrulesep 
    \begingroup 
      \color{#1}
      \hrule height\belowrulesep 
    \endgroup 
  }
} 
\newcommand*{\aboverulesepcolor}[1]{
  \noalign{
    \begingroup 
      \color{#1} 
      \hrule height\aboverulesep 
    \endgroup 
    \kern-\aboverulesep 
  }
}

\newcommand{\myparagraph}[1]{\paragraph{\textbf{\textup{#1.}}}}

\newcommand{\maxx}[2]{\bigvee\limits_{#1}^{#2}}

\newcommand{\layerout}[1][]{
  \ifthenelse{\isempty{#1}}
    {g}
    {g\left(#1\right)}
}
\newcommand{\layerin}[1][]{
  \ifthenelse{\isempty{#1}}
    {f}
    {f\left(#1\right)}
}
\newcommand{\kernel}[1][]{
  \ifthenelse{\isempty{#1}}
    {h}
    {h\left(#1\right)}
}

\def\eg{\emph{e.g. }}
\def\ie{\emph{i.e. }}

\begin{document}
\maketitle
\begin{abstract}
Signals from different modalities each have their own combination algebra which affects their sampling processing.
RGB is mostly linear; depth is a geometric signal following the operations of mathematical morphology.
If a network obtaining RGB-D input has both kinds of operators available in its layers, it should be able to give effective output with fewer parameters.
In this paper, morphological elements in conjunction with more familiar linear modules are used to construct a mixed linear-morphological network called \textbf{HaarNet}.
This is the first large-scale linear-morphological hybrid, evaluated on a set of sizeable real-world datasets.
In the network, morphological Haar sampling is applied to both feature channels in several layers, which splits extreme values and high-frequency information such that both can be processed to improve both modalities. 
Moreover, morphologically parameterised ReLU is used, and morphologically-sound up-sampling is applied to obtain a full-resolution output.
Experiments show that HaarNet is competitive with a state-of-the-art CNN, implying that morphological networks are a promising research direction for geometry-based learning tasks.
\keywords{Machine learning  \and Morphological neural networks \and Multi-modal learning.}
\end{abstract}

\section{Introduction} \label{sec:introduction}
Semantic segmentation is a challenging pixel-level prediction task in computer vision.
It has become common practice \cite{silberman2012indoor,gupta2014learning} to use multi-modal \emph{colour} and \emph{depth} (RGD-D) data since failure modes arising from one modality (\eg homogeneity, noise, sparsity, etc.) can often be avoided when relying on multiple modalities.
Datasets such as \cite{silberman2012indoor,armeni2017joint,cordts2016cityscapes} facilitate research that delves into combining features from different signals by providing data from different sensors.
However, optimal fusion strategies for the colour and depth data modalities are still an open area of research.

Through the works of Serra \cite{serra1982image}, the computer vision community is familiarised with the underlying algebraic structure of data that is acquired using probing contact (\eg LiDAR and radar) \cite{ritter1996introduction,sussner1998morphological,serra1992overview}.
This field is called \emph{mathematical morphology}, and it differs from the algebra of linear diffusion that is used to build convolutional neural networks (CNNs).
Recently, it was shown that morphological operations are suitable for processing specific types of data, such as depth data \cite{groenendijk2022morphpool}.
Besides the benefits for encoding input signals, morphological sampling and interpolation may help to predict sharply delineated semantic boundaries.
While there exist examples of methods that explore morphological networks \cite{mondal2022morphological}, or embed morphological operations in few-layer linear networks \cite{shen2019deep,franchi2020deep,dimitriadis2020advances,hernandez2020hybrid,roy2021morphological,velasco-forero2022fixed}, large-scale linear-morphological hybrids have not received much research attention.

This paper proposes the construction of a mixed linear-morphological network, coined HaarNet, to exploit linear and morphological operations jointly. 
The proposed morphological modules are: a down-sampling operation based on the morphological Haar transform; non-linearities formulated as parameterised dilations; and an up-sampling procedure that can innately delineate semantic boundaries. 

\section{Related Work} \label{sec:related-work}
\begin{figure}[t]
\centering
\includegraphics[width=0.9\textwidth]{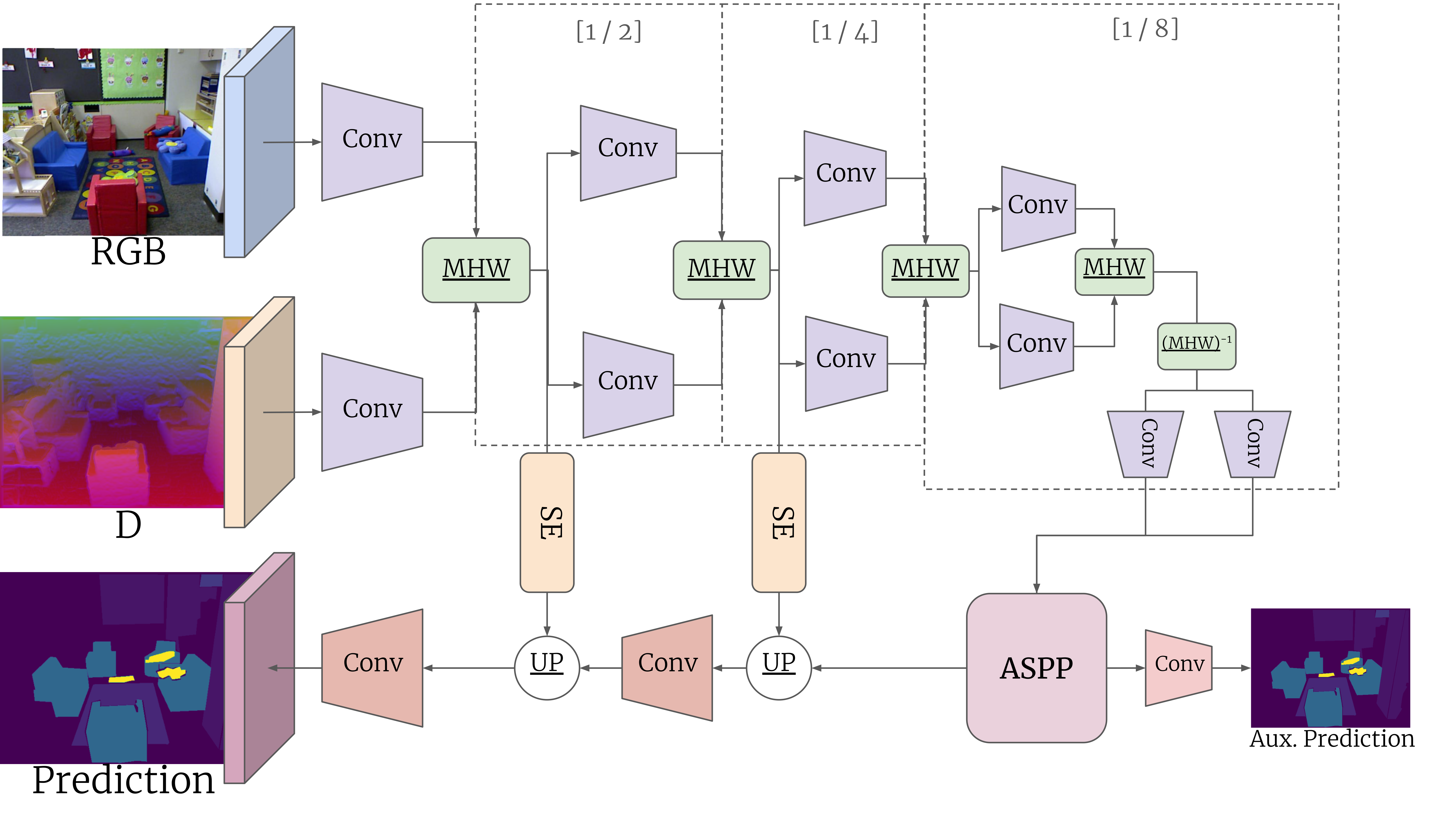}
\caption{
\textbf{The Morphological HaarNet.}
Depth and RGB modalities are separately encoded, but during down-sampling the Morphological Haar Wavelet (\underline{MHW}, see~\autoref{fig:haar-wavelet-block} for more details) blocks ensure that the down-sampled signal is improved using high-frequency details from both modality feature signals.
Up-sampling happens through morphological up-sampling and skip connections (\underline{UP}).
Both morphological types are underlined in this overview.
The skip connections receive channel-wise attention from (multi-modal) SE \cite{hu2018squeeze} blocks.
Multi-resolution features are combined using ASPP \cite{chen2017deeplab}.
All non-linearities are replaced by the morphological dilation.
}
\label{fig:haarnet-overview}
\end{figure}
Fusion of RGB and depth images in neural networks may be done at the input or output stage, although the most common method is ``middle'' fusion.
Middle fusion can be either operator-based or attention-based. 
Attention-based fusion ensures that long-range dependencies between modalities can be modelled, rather than simply fusing on local modality-based features as operator-based fusion would do.
While there are many articles addressing this type of fusion, of note is \cite{chen2020bi}, who proposes to re-calibrate both feature encodings using Separation-and-Aggregation (SAGate) at each down-sampling step in the encoder. 
Moreover, a propagation module is introduced that normalises information flow, allowing the authors to make better use of pre-trained weights.
In \cite{chen2021global}, the authors observe that alignment of features from RGB and depth is not often addressed, and propose the Global-Local Propagation network that makes use of attention blocks to warp and align feature volumes.
CANet \cite{zhou2022canet} uses a combination of channel-wise and spatial attention mechanisms to fuse depth and RGB using separate pathways.

The method proposed in the current paper, too, considers the attention-by-sampling, but in a more structured way through the morphological Haar transform.
Most similar to our proposed method is the MSFNet introduced by \cite{jiang2022multi}.
In MSFNet, a \emph{linear} Haar Wavelet transform is used to decompose high resolution signals into constituents, ensuring high frequency information is preserved throughout down-sampling. 
However, instead of using the wavelets as a means of down-sampling, the inverse transform is used to reconstruct the feature signal at the original resolution. 
In addition, the depth features are only used to attend the RGB features, so no joint optimisation of feature representations is being performed.
HaarNet, on the other hand, allows to freely optimise the set of features from both modalities.

\section{Method} \label{sec:method}
To specify the composition of a Morphological Haar Wavelet Network, a brief overview of the fundamental morphological operations is required:
the morphological \emph{dilation} is defined on the semi-ring $\left\{\mathbb{R}_{-\infty}, \bigvee, + \right\}$ where $\bigvee$ denotes the supremum operation and $+$ is addition. 
This algebraic system extends the set of real numbers $\mathbb{R}$ with minus infinity: $\mathbb{R}_{-\infty} \equiv \mathbb{R} \cup -\infty$ \cite{ritter1996introduction}.
The morphological \emph{erosion} is the dual of the morphological dilation, and is constructed with infimum $\bigwedge$ over the extended set of reals $\{\mathbb{R} \cup \infty\}$.
The dilation and erosion can be parameterised by means of a structuring element $h$.
The dilation is defined as
\begin{equation} \label{eq:morphological-dilation}
    \layerout[\mathbf{x}] = \left[\layerin \boxplus \kernel\right](\mathbf{x}) = \maxx{\mathbf{z}}{} \layerin[\mathbf{x} - \mathbf{z}] + \kernel[\mathbf{z}]\,,
\end{equation}
where $\layerin, \layerout$ are the inputs and outputs of a layer in a neural network respectively, and $\mathbf{x}, \mathbf{z}$ are taken from indicator sets.
Unlike convolutions in CNNs, morphological operations are often only applied in a channel-wise manner \cite{groenendijk2022morphpool}; back-propagation over the maximum element results in slow and complicated learning \cite{groenendijk2022backprop}.
Note that dilation is algebraically similar to convolution, though in the \emph{logarithmic} sense \cite{burgeth2005logarithmic}.
As a consequence, it can serve as a signal processing framework with an essentially analogous structure to linear signal processing.

There are four novel elements on which the morphological HaarNet is built:
\begin{itemize}
    \item The morphological Haar Wavelet module, which improves down-sampling compared to common strategies like strided convolutions or pooling.
    \item The ReLU non-linearity that is shown to be a special case of the morphological dilation; it can be parameterised to be a learnable threshold-filter.
    \item An improved sampling procedure relative to \cite{groenendijk2022morphpool}, following seminal papers on sampling theory in morphology \cite{heijmans1991morphological,heijmans2000nonlinear} rather than being provenance-based.
    \item The use of additional pre-training to compensate for the reduced effectiveness of transfer learning when compared to baselines similar to a pre-trained ResNet \cite{he2016deep}.
\end{itemize}

The full network overview is shown in~\autoref{fig:haarnet-overview}, which details at which stages multi-modal features are fused.
Note that the network uses Squeeze-and-Excitation blocks \cite{hu2018squeeze} for attending channel-wise information during skip connections, although they are modified to attend separately to each modality feature volume before feature fusion.
Another module that is used, and that is prevalent in pixel-level semantic prediction networks, is the multi-scale feature pyramid ASPP \cite{chen2017deeplab} to improve predictive performance for long-range semantic dependencies.
These blocks are naturally augmented by morphological non-linearities (see~\hyperref[sec:method:relu-as-dilation]{Section 3.2}), to truly make this network a linear-morphological hybrid.

\subsection{Morphological Haar Wavelets} \label{sec:method:haar-module}
\begin{figure}[t]
\centering
\includegraphics[width=0.7\textwidth]{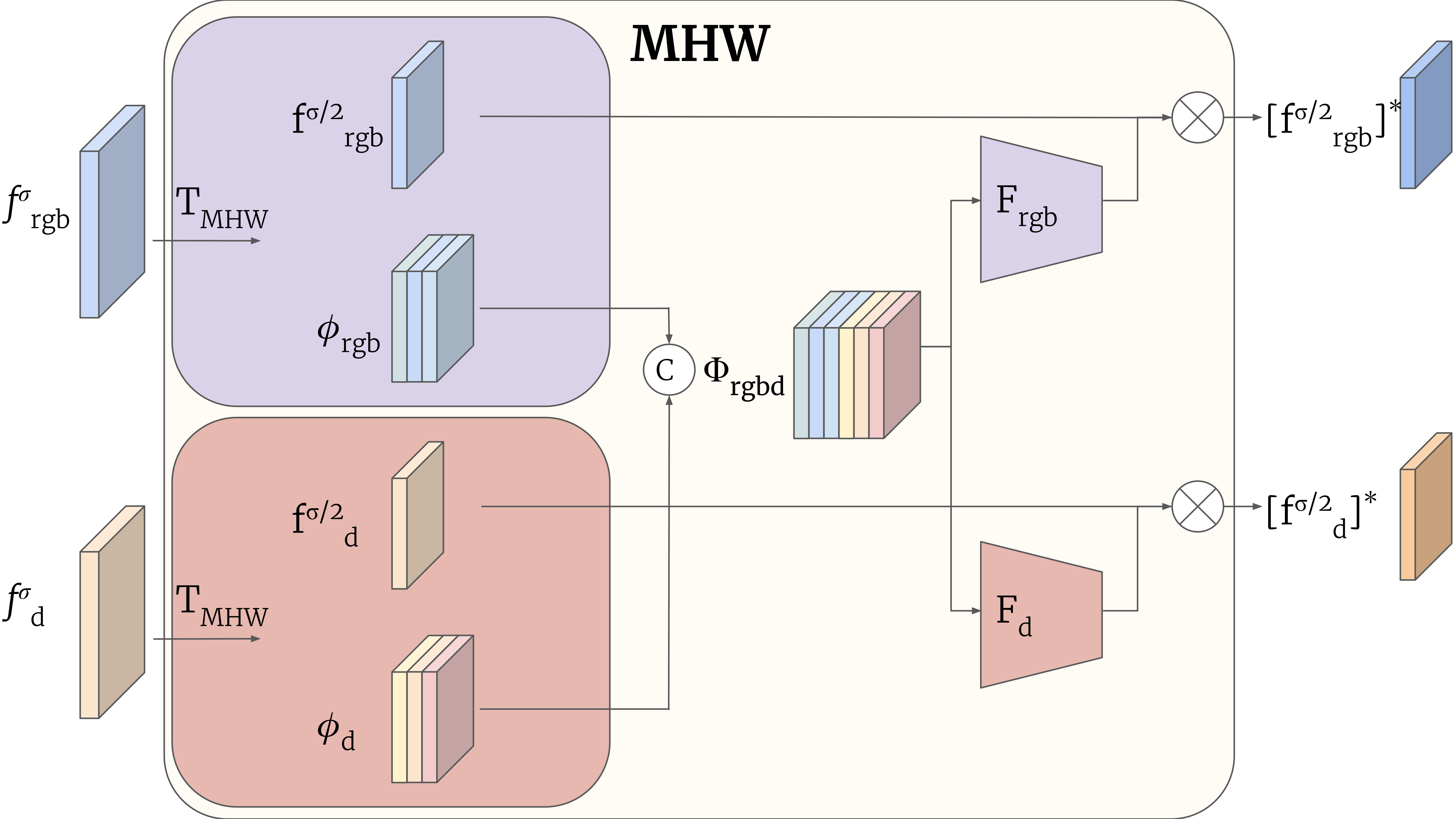}
\caption{
\textbf{The Morphological Haar Wavelet} (MHW) block, which down-samples the signal $f^{\sigma}$ by decomposing into subbands $f^{\sigma/2}$ and $\phi$. 
A 2-layer MLP $F$ is used to attend to channel-wise information based on combined high-frequency details $\Phi_{rgbd}$ for both modalities.
This yields updated feature representations $\left[f_{rgb}^{\sigma/2}\right]^{*}$ and $\left[f_{d}^{\sigma/2}\right]^{*}$.
}
\label{fig:haar-wavelet-block}
\end{figure}
The morphological Haar wavelet transform decomposes signal $f^{\sigma}\colon \mathbb{R}^{H\times W} \to \mathbb{R}$ at a particular scale $\sigma$ into four subband signals: Down-sampled $f^{\sigma/2}\colon \mathbb{R}^{H/2\times W/2} \to \mathbb{R}$ and vertical, horizontal, and diagonal details $\phi^{v}_{f},\phi^{h}_{f},\phi^{d}_{f}\colon \mathbb{R}^{H/2\times W/2} \to \mathbb{R}$ respectively.
A particularly helpful property of morphological wavelets over linear wavelets is that they retain local maxima of the signal over multiple scales; similar to pooling, this implies the network achieves invariance to particular kinds of noise \cite{jarrett2009best}.
The morphological Haar wavelet transform \cite{heijmans1991morphological} is defined
\begin{subequations}
\setlength{\arraycolsep}{4.5pt}
    \begin{align}
        \begin{split}
            f^{\sigma/2} &= f^{\sigma} \boxplus \begin{bmatrix*}[r] 0 & 0 \\ 0 & 0 \end{bmatrix*} \,,
        \end{split} \\
        \begin{split}
            \phi_{f} &\equiv \left\{\phi^{v}_{f},\phi^{h}_{f},\phi^{d}_{f}\right\}  = f^{\sigma} *\left\{ \begin{bmatrix*}[r] -1 & -1 \\ 1 & 1 \end{bmatrix*},  \begin{bmatrix*}[r] -1 & 1 \\ -1 & 1 \end{bmatrix*}, \begin{bmatrix*}[r] 1 & -1 \\ -1 & 1 \end{bmatrix*}\right\} \,,
        \end{split}
    \end{align}
\end{subequations}
where both morphological dilation $\boxplus$ and convolution $*$ are applied at a stride of 2 to get non-overlapping windows, and to ensure that the spatial dimensions of the signal $f^{\sigma}$ are reduced by a factor 2.
Clearly, $f^{\sigma/2}$ is similar to the application of max pooling which computes the local maximum value in a 2-by-2 neighbourhood.
Max pool and $f^{\sigma/2}$ both reduce spatial feature size in addition to obtaining maximum-amplitude coefficients \cite{chen2013deep,xie2014spatial}.
On top of this, detail signals $\phi$ encode all directional high-frequency information that was lost during down-sampling.
A property of the wavelet transform is that it is possible to fully restore $f$ by the inverse transform using subbands $f^{\sigma/2}$ and $\phi_f$.

The idea of the Morphological Haar Wavelet (MHW) block is to use combined directional details of RGB and depth modalities $\Phi_{rgbd} \equiv \left\{\phi_{rgb}, \phi_{d}\right\}$ to simultaneously improve both the down-sampled RGB feature signal $f^{\sigma/2}_{rgb}$ and the depth feature signal $f^{\sigma/2}_{d}$.
To this end, MHW uses two distinct 2-layer CNNs $F_{rgb}, F_{d}$.
Each $F$ takes as input the combined high-frequency signal $\Phi_{rgbd}$ to obtain the updated and improved down-sampled signals $[f^{\sigma/2}]^{*}$. 
For RGB features this can be expressed as
\begin{equation}
    \left[f_{rgb}^{\sigma/2}\right]^{*}(\mathbf{x}) = f^{\sigma/2}_{rgb} (\mathbf{x}) \otimes \sigma\left(F_{rgb}\left(\Phi_{rgbd}\left(\mathbf{x}\right)\right)\right) \,,
\end{equation}
where $\sigma(\cdot)$ is a sigmoid, $\otimes$ is element-wise multiplication, and $\mathbf{x}$ taken from the indicator set over the signal. 
Similarly, for the improved depth features $[f_{d}^{\sigma/2}]^{*}$. 
The combined operation ensures not only that high frequency information is used explicitly during down-sampling, but also forces the resulting information to be dependent on both modalities through their common $\Phi_{rgbd}$.
The full procedure is depicted schematically in~\autoref{fig:haar-wavelet-block}.

\subsection{Formulating ReLU as Dilation} \label{sec:method:relu-as-dilation}
The ReLU activation layer \cite{fukushima1975cognitron} is a special case of the morphological dilation with a flat structuring element that is the morphological delta-function, and a morphological bias term.
The morphological delta-function $\delta(\cdot)$ is 0 only at location 0, and $-\infty$ elsewhere.
To demonstrate the equivalence between ReLU and morphological dilation, first consider the definition of the ReLU:
\begin{equation}
    \xi(\mathbf{x}) =     
    \begin{cases}
    \xi(\mathbf{x})     &\text{if}\quad \xi(\mathbf{x}) \geq 0 \\
    0                   &\text{otherwise.}
    \end{cases}
\end{equation} \label{eq:morphological-dilation-with-bias}
Similar to the convolutional bias, the dilation can be written to include a morphological bias term \cite{charisopoulos2017morphological}, denoted $h_{0}$.
Taking the above into account, the following can be obtained
\begin{align}
    \xi\left(\layerin[\mathbf{x}]\right) &= 0 \vee \layerin[\mathbf{x}] \,,&\\
    &= 0 \vee \left[\layerin[\mathbf{x}] \boxplus \delta(\mathbf{x})\right] &\left(\text{using }\delta(\cdot)\right)\,, \\
    &= \kernel_{0} \vee \left[\layerin \boxplus \kernel\right](\mathbf{x}) &\text{(Setting $\kernel_{0}$ = 0)}\,.
\end{align}
Thus, a form is reached that suggests a natural extension of ReLU by morphological kernel $\kernel$ and bias $h_{0}$.
In HaarNet, all activation layers are replaced by the morphological dilation in which the morphological bias term $\kernel_{0}$ is learnable.
Such a learnable ReLU was coined Flexible ReLU (FReLU) in \cite{qiu2018frelu}, but the connection to the morphological dilation was not recognised in this work.
A strong belief is held by the authors that understanding of networks improves by recognising that nearly all modern CNNs exploit inherently morphological layers through ReLU \cite{velasco2022morphoactivation} and pooling \cite{groenendijk2022morphpool}.

\subsection{Morphological Up-sampling}
In encoder-decoder architectures, up-sampling from low-resolution features is vital to obtain high-quality predictions.
In \cite{groenendijk2022morphpool}, it was shown that non-linear up-sampling can aid semantic networks, since it delineate boundaries better than linear up-sampling.
The procedure also did not introduce sparsity the way standard unpooling does \cite{zeiler2011adaptive,zeiler2014visualizing}.
However, morphological up-sampling can be improved using insights from \cite{heijmans1991morphological}, where it is proven by the theory of adjunctions that the down-sampling dilation has an up-sampling dual that is an erosion with the same structuring element.
In addition, up-sampling is of equidistant nature: intervals at which to place back values are kept constant.
As a consequence, the total reconstruction operator $\rho$ is a closing: $\rho[f] = \left[f \boxplus \kernel\right] \boxminus \kernel$.
Note, however, that in prediction tasks \emph{reconstruction} of the same signal is not the goal, rather the input signal must be \emph{processed} for analysis.
Unlike \cite{heijmans1991morphological}, this paper is concerned with obtaining the most expressive sampling processing operator.

Based on experimentation, the proposed morphological up-sampling operator for prediction networks is as follows:
First, equidistant sampling is used, and not provenance mapping as in \cite{groenendijk2022morphpool}. This means low-resolution features are placed back at a regular interval.
Second, all sampling operators are dilations. 
Third, structuring elements are not constrained to be the same across down- and up-sampling operators. That is, they are all parameterised individually.
Lastly, in HaarNet, the up-sampling procedure is combined with a skip connection at the same resolution.

\section{Experiments} \label{sec:experiments}
\begin{table}[t]
    \centering
    \caption{\textbf{Effects of Pre-trained Weights.}
    ResNet50 encoders are initialised either randomly or use weights pre-trained from ImageNet.
    HaarNet performs better on two out of three metrics when randomly initialised. 
    When ImageNet weights are used, a basic ResNet benefits most. 
    }
    \resizebox{0.6\textwidth}{!}{
        \begin{tabular}{@{\extracolsep{4pt}}lccc@{}}
        \toprule
        & mIoU & pixel accuracy & Boundary F1 \\
        \cline{2-4}
        \multicolumn{4}{a}{Random Initialisation} \\
        Middle Fusion Dual ResNet       & 0.193 & 0.582 & 0.220 \\
        SA-Gate \cite{chen2020bi}       & \bf0.266 & 0.603 & 0.213 \\
        HaarNet                         & 0.252 & \bf0.624 & \bf0.231 \\
        \cline{1-4}
        \multicolumn{4}{a}{Naive Transfer of ImageNet Weights} \\
        Middle Fusion Dual ResNet       & 0.382 & 0.687 & \bf0.314 \\
        SA-Gate \cite{chen2020bi}       & \bf0.408 & \bf0.697 & 0.267 \\
        HaarNet                         & 0.360 & 0.687 & 0.292 \\
        \bottomrule
        \end{tabular}
    }
    \label{tab:pretraining-results}
\end{table}
\begin{table}[t!]
    \centering
    \caption{\textbf{Indoor Segmentation Results.}
    When trained using the same set-up, HaarNet performs at least on par with the baselines using the least amount of additional parameters, as reported in the second column.
    }
    \resizebox{\textwidth}{!}{
        \begin{tabular}{@{\extracolsep{4pt}}lc|ccc|ccc@{}}
        \toprule
        Network && \multicolumn{3}{c}{NYUv2} & \multicolumn{3}{c}{2D-3D-S} \\
        &+\#param & mIoU & pixel accuracy & Boundary F1 & mIoU & pixel accuracy & Boundary F1 \\
        \cline{3-5} \cline{6-8}
        Dual ResNet Baseline                    & +25.8M & 0.474 & 0.749 & 0.373 & 0.565 & 0.787 & 0.564 \\
        SAGate \cite{chen2020bi}                & +21.8M & 0.504 & 0.768 & 0.381 & 0.561 & \bf0.806 & \bf0.577 \\
        HaarNet                                 & +13.1M & \bf0.507 & \bf0.770 & \bf\underline{0.392} & \bf0.568 & 0.794 & 0.567 \\
        \bottomrule
        \end{tabular}
    }
    \label{tab:indoor-results}
\end{table}
In this section, HaarNet is applied to a semantic segmentation task using both RGB and depth modalities on three different datasets, both indoor and outdoor.
HaarNet is compared to state-of-the-art baselines, most notably SAGate \cite{chen2020bi} as it is encoder-decoder-based, reaches state-of-the-art consistently across datasets, and --most importantly-- public code is available for replication of the results and for a fair comparison.
Large-scale linear-morphological hybrids have not received research attention; the goal of the experiments is to show that even when network design principles from linear networks are used, linear-morphological hybrids still perform at least on par with linear counterparts.

\myparagraph{Implementation}
All modules are implemented in PyTorch \cite{paszke2019pytorch}, except for morphological operations, which are absent in PyTorch and potentially intractable when naively done. 
These operations are implemented directly in CUDA/C++.

\myparagraph{Dataset \& Training}
The experiments are performed on three datasets: NYUv2 ($N_{\text{train}}$=795, $N_{\text{test}}$=654, $C$=40), 2D-3D-S ($N_{\text{train}}$=52903, $N_{\text{test}}$=17593, $C$=13), and CityScapes ($N_{\text{train}}$=2975, $N_{\text{test}}$=500, $C_{train}$=30, $C_{test}$=19).
The first two datasets are indoor datasets, the last is an outdoor autonomous driving dataset.
RGB images are normalised with training set statistics to follow a zero-mean Gaussian.
According to convention, depth images are converted to the HHA encoding \cite{gupta2014learning} and normalised in the range $\left[0,1\right]$.
Based on initial experimentation, all networks are trained using SGD with Nesterov Momentum \cite{sutskever2013importance} with a learning rate $\lambda_{0}$ of $5\mathrm{e}{-3}$.
The learning rate is updated at each epoch by a polynomial scheduler according to $\lambda_{epoch} = \lambda_{0} \left(1 - \tfrac{epoch}{500}\right)^{0.9}$.
The networks are trained for 500, 12, and 200 epochs for the NYU, 2D-3D-S, and CityScapes datasets respectively.
Models are evaluated using mean Intersection over Union (mIoU), pixel accuracy, and a boundary F1-score \cite{csurka2004good} to measure performance at semantic edges.

\subsection{The Necessity of Pre-training} \label{sec:experiments:pre-training}
It is hypothesised that many contemporary networks are architecturally biased to the pre-trained ResNet weights on ImageNet, since pre-training is such a powerful means to obtain state-of-the-art performance \cite{yosinski2014transferable}.
Specifically, when the architecture starts to significantly differ from that of the ResNet, exploiting the pre-trained weights becomes less valuable.
When a fundamentally different network, like HaarNet, is introduced, it should undergo pre-training for fair assessment of its novel capabilities.

Pre-training new encoders on ImageNet, however, is not computationally feasible in most scenarios because of the scale of this dataset.
To adapt, this paper joins the three datasets (\ie NYUv2, 2D-3D-S, and CityScapes) and maps all semantic labels to a joint set of 67 labels, including semantic classes from both indoor and outdoor scenes.
Pre-training is performed in two phases: first, the decoder is frozen and the encoder is trained using the joint dataset. 
Second, the decoder is trained on a specific dataset with a frozen encoder.

Consider~\autoref{tab:pretraining-results}, in which the pre-training hypothesis is tested.
All networks use a ResNet50 encoder, although both the SAGate baseline and HaarNet make significant changes to their encoder.
SAGate compensates for architectural changes by the normalising the outputs of their fusion blocks; HaarNet has no such inherent rectification.
It is shown that without any pre-training, HaarNet performs best on two out of three metrics.
However, when ImageNet weights for ResNet are used as a pre-training procedure, HaarNet benefits least; a basic ResNet benefits most.

\subsection{Segmentation Performance} \label{sec:experiments:indoor}
\begin{figure}[h!]
    \begin{tabular}{@{\hspace{0cm}}c@{\hspace{0cm}}@{\hspace{0cm}}c@{\hspace{0cm}}@{\hspace{0cm}}c@{\hspace{0cm}}@{\hspace{0cm}}c@{\hspace{0cm}}@{\hspace{0cm}}c@{\hspace{0cm}}@{\hspace{0cm}}c@{\hspace{0cm}}}
    \toprule  \vspace*{-2.25em}
    \centering
    & & & & & \\
    \multicolumn{2}{l}{\subfloat{\includegraphics[width=0.3\textwidth,clip]{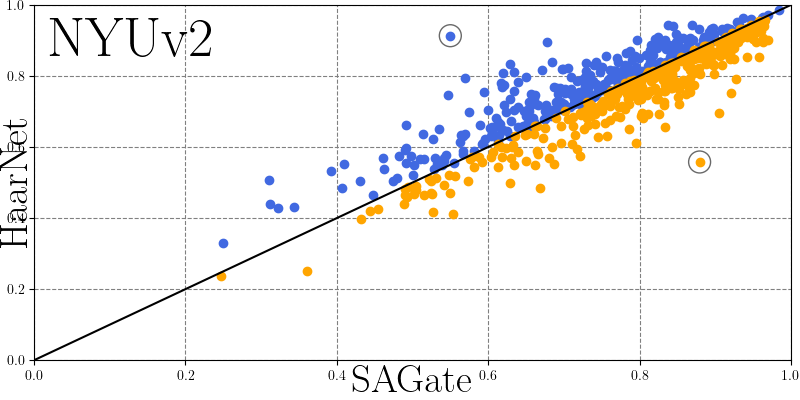}}} &
    \multicolumn{2}{c}{\subfloat{\includegraphics[width=0.3\textwidth,clip]{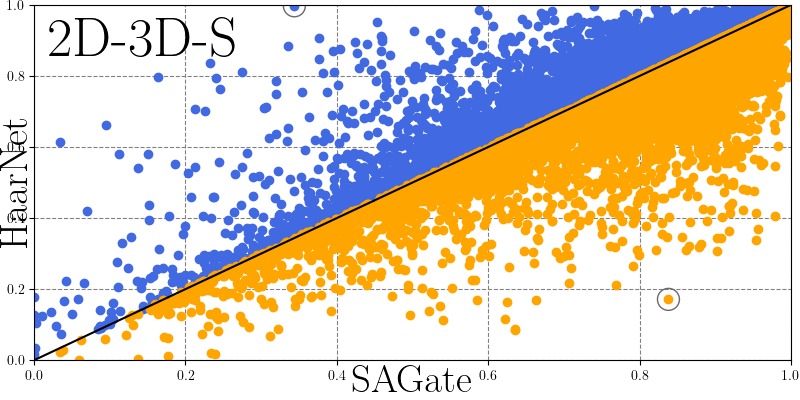}}} &
    \multicolumn{2}{r}{\subfloat{\includegraphics[width=0.3\textwidth,clip]{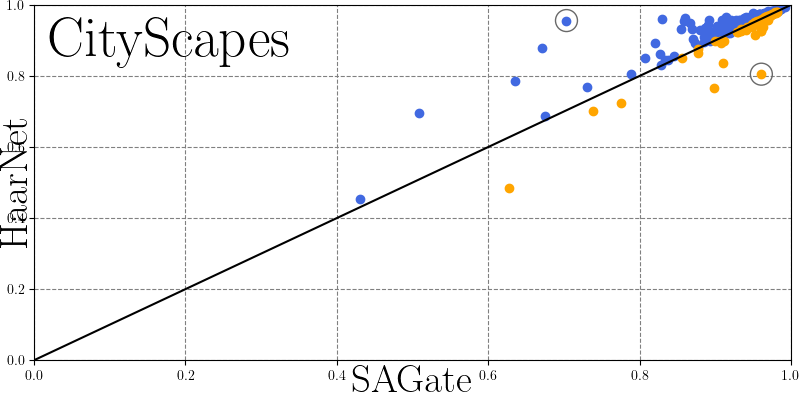}}} \vspace*{-0.5em} \\
    \midrule
    RGB & HHA & Ground Truth & Baseline & SAGate \cite{chen2020bi} & HaarNet \vspace*{-1.25em} \\
    \subfloat{\includegraphics[width=0.16\textwidth,clip]{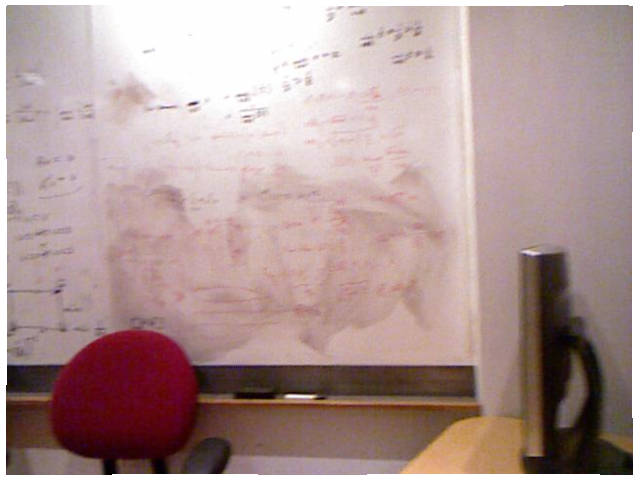}} &
    \subfloat{\includegraphics[width=0.16\textwidth,clip]{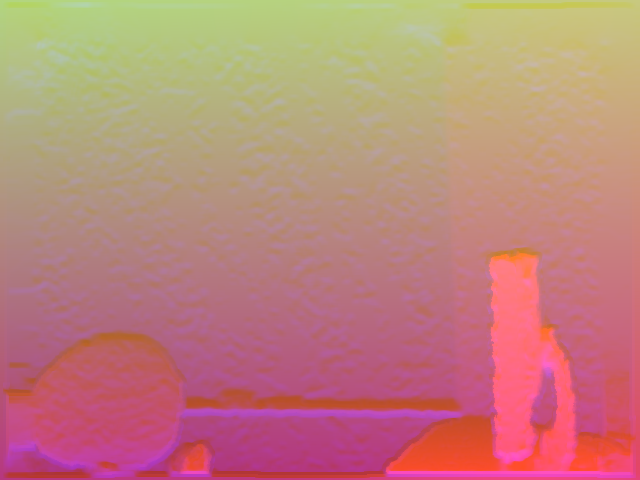}} &
    \subfloat{\includegraphics[width=0.16\textwidth,clip]{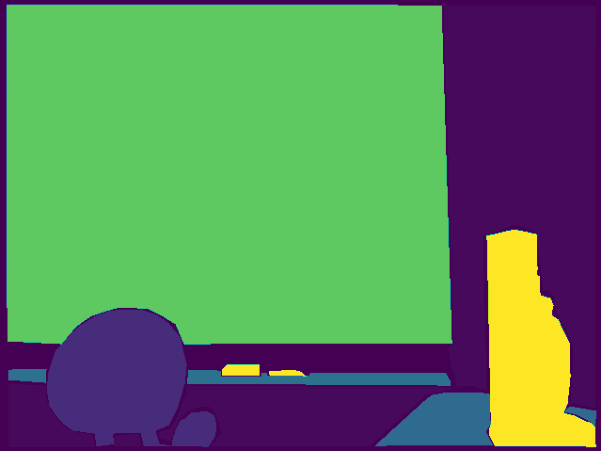}} &
    \subfloat{\includegraphics[width=0.16\textwidth,clip]{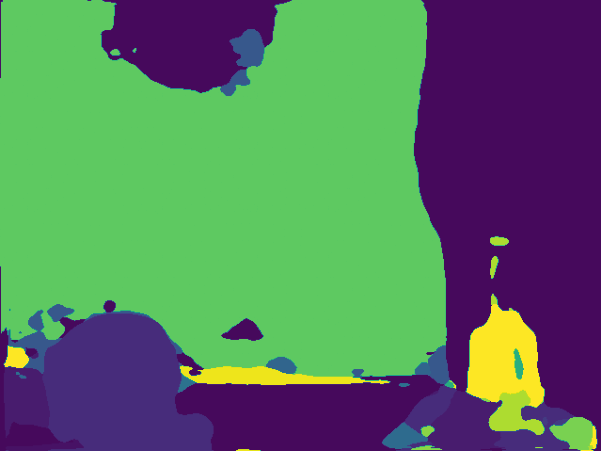}} &
    \subfloat{\includegraphics[width=0.16\textwidth,clip]{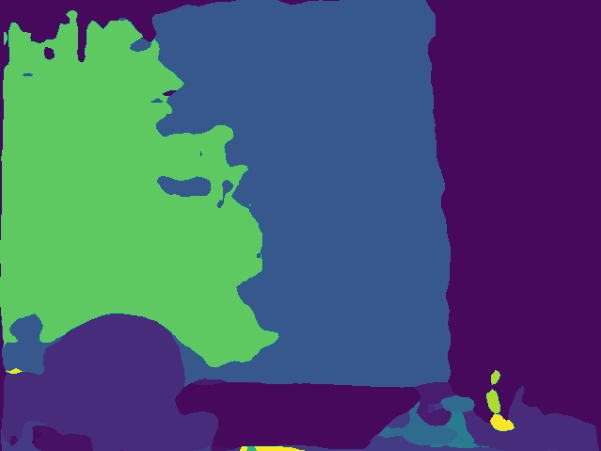}} & 
    \subfloat{\includegraphics[width=0.16\textwidth,clip]{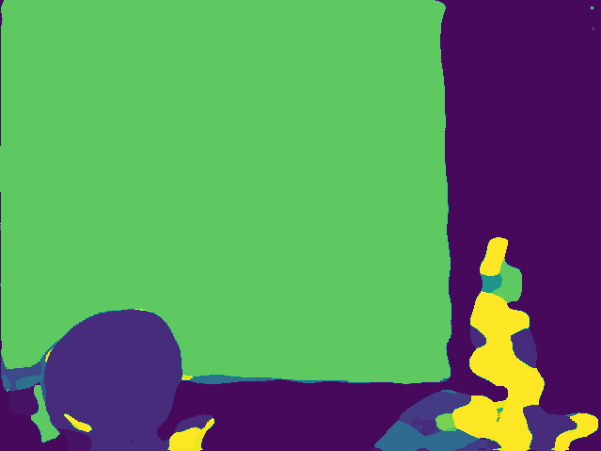}} \vspace*{-1.25em} \\
    \subfloat{\includegraphics[width=0.16\textwidth,clip]{comparison/nyu_rgb_worst.png}} &
    \subfloat{\includegraphics[width=0.16\textwidth,clip]{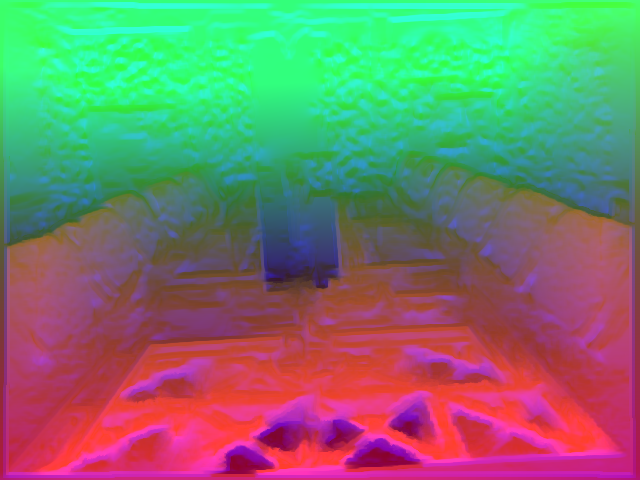}} &
    \subfloat{\includegraphics[width=0.16\textwidth,clip]{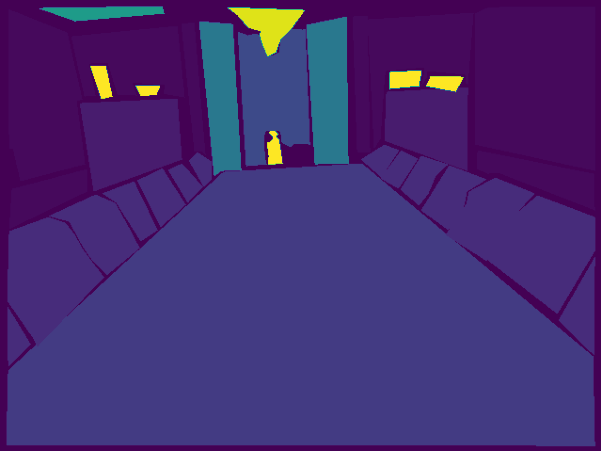}} &
    \subfloat{\includegraphics[width=0.16\textwidth,clip]{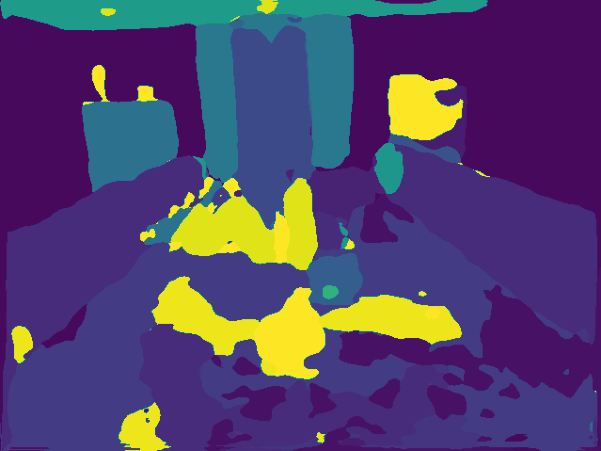}} &
    \subfloat{\includegraphics[width=0.16\textwidth,clip]{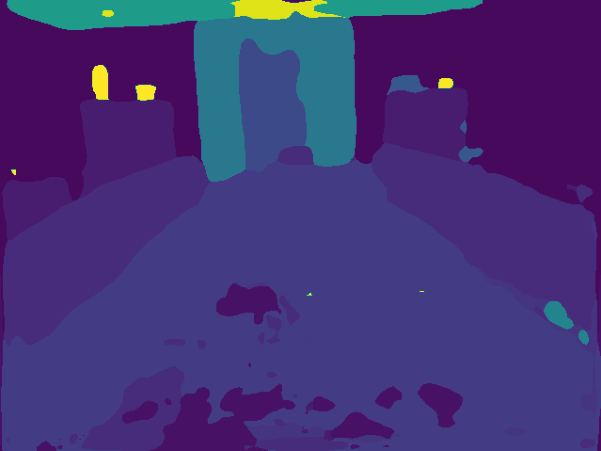}} & 
    \subfloat{\includegraphics[width=0.16\textwidth,clip]{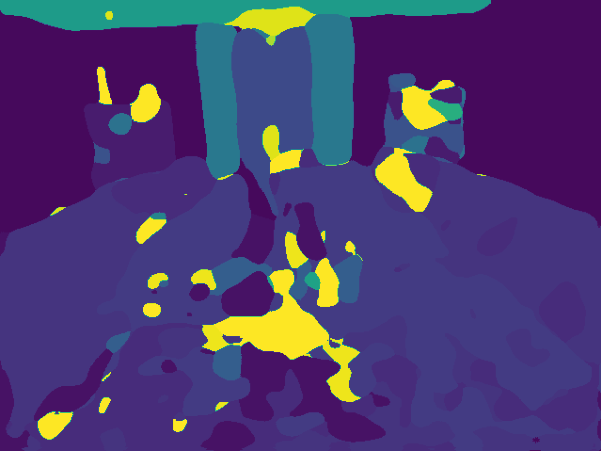}} \vspace*{-0.45em} \\
    \midrule
    & & & & & \vspace*{-2.35em} \\
    \subfloat{\includegraphics[width=0.16\textwidth,clip]{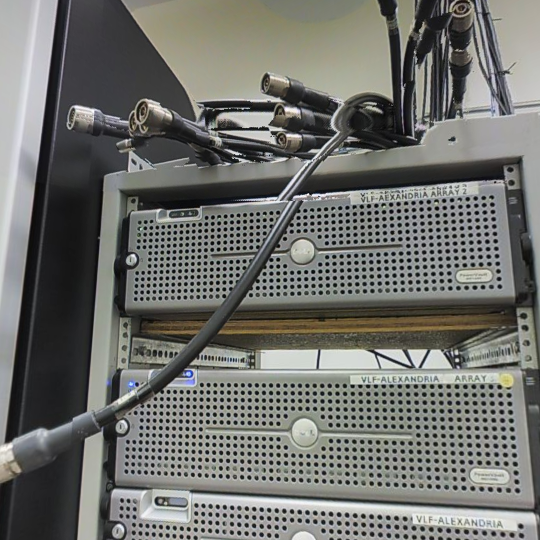}} &
    \subfloat{\includegraphics[width=0.16\textwidth,clip]{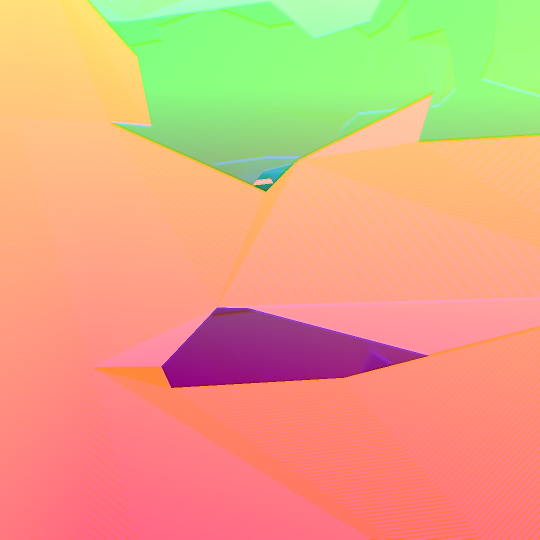}} &
    \subfloat{\includegraphics[width=0.16\textwidth,clip]{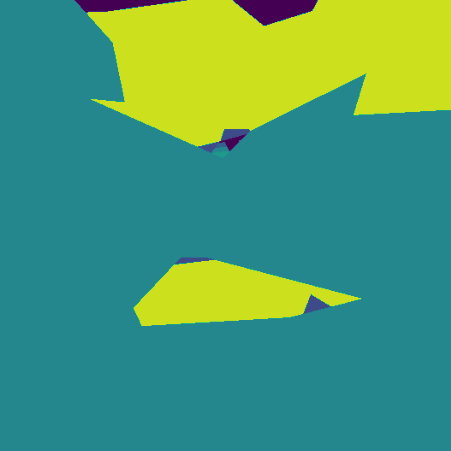}} &
    \subfloat{\includegraphics[width=0.16\textwidth,clip]{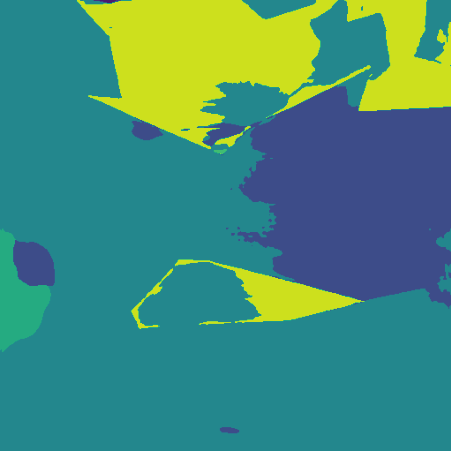}} &
    \subfloat{\includegraphics[width=0.16\textwidth,clip]{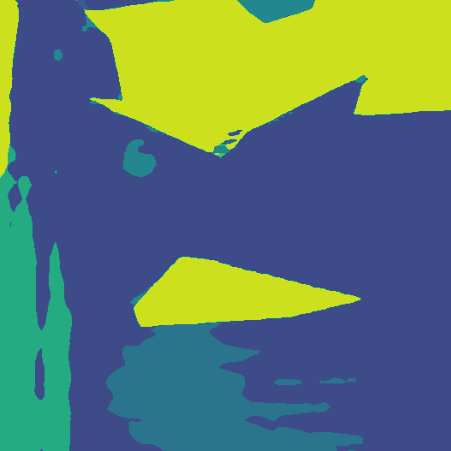}} & 
    \subfloat{\includegraphics[width=0.16\textwidth,clip]{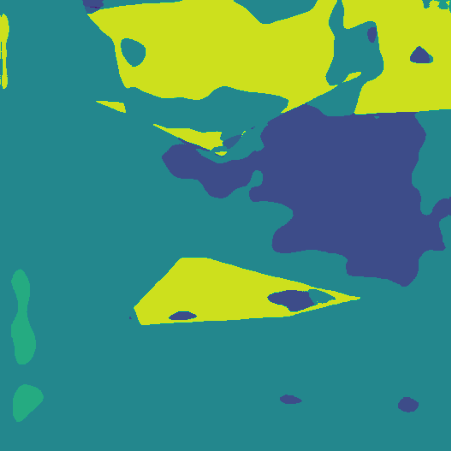}} \vspace*{-1.25em} \\
    \subfloat{\includegraphics[width=0.16\textwidth,clip]{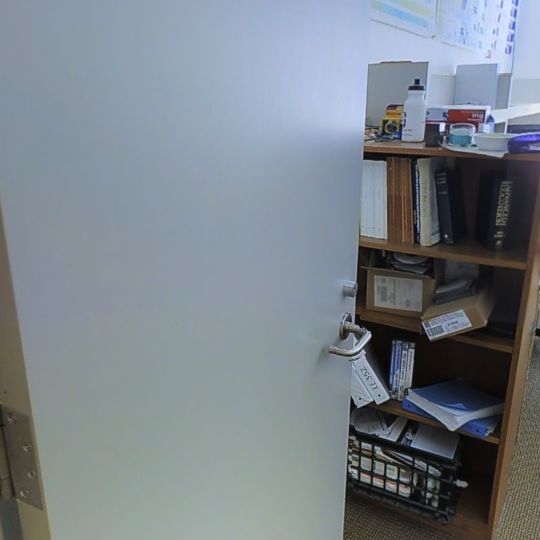}} &
    \subfloat{\includegraphics[width=0.16\textwidth,clip]{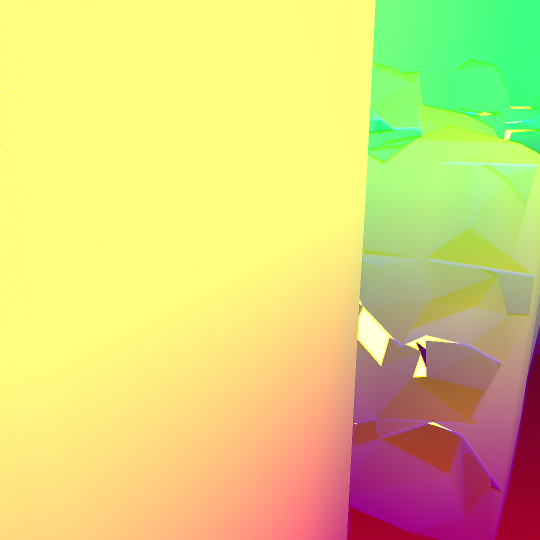}} &
    \subfloat{\includegraphics[width=0.16\textwidth,clip]{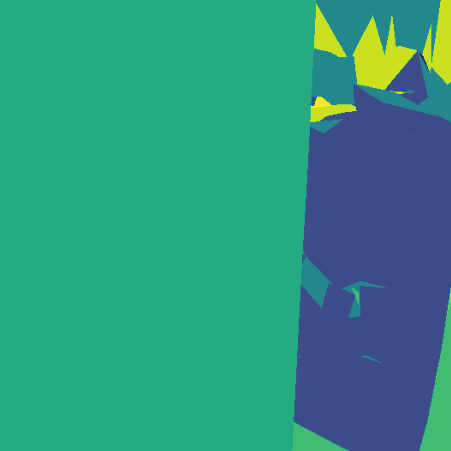}} &
    \subfloat{\includegraphics[width=0.16\textwidth,clip]{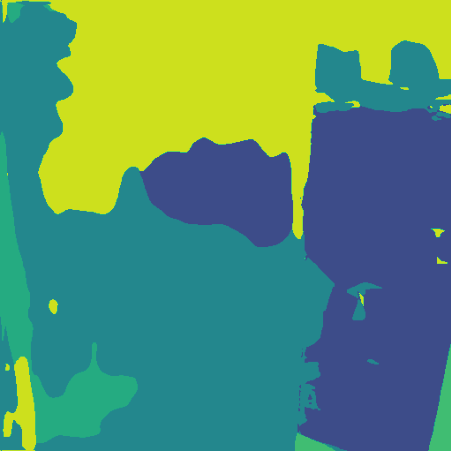}} &
    \subfloat{\includegraphics[width=0.16\textwidth,clip]{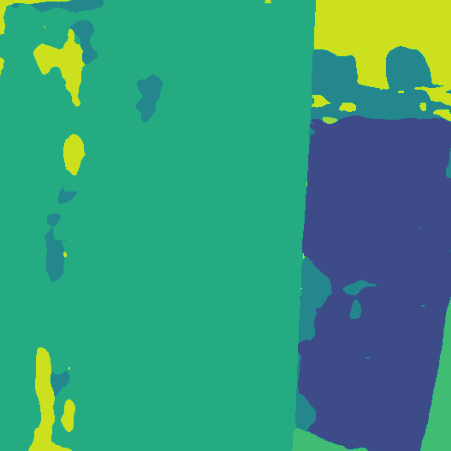}} & 
    \subfloat{\includegraphics[width=0.16\textwidth,clip]{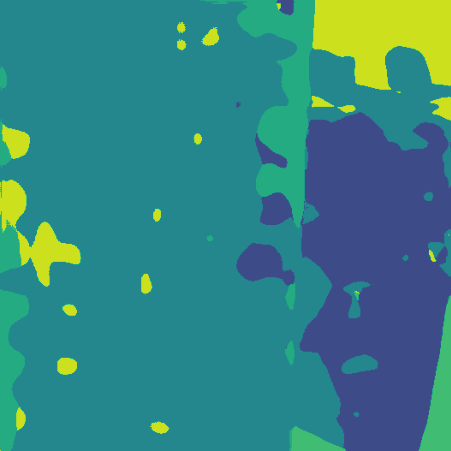}} \vspace*{-0.45em} \\
    \midrule
    & & & & & \vspace*{-2.35em} \\
    \subfloat{\includegraphics[width=0.16\textwidth,clip]{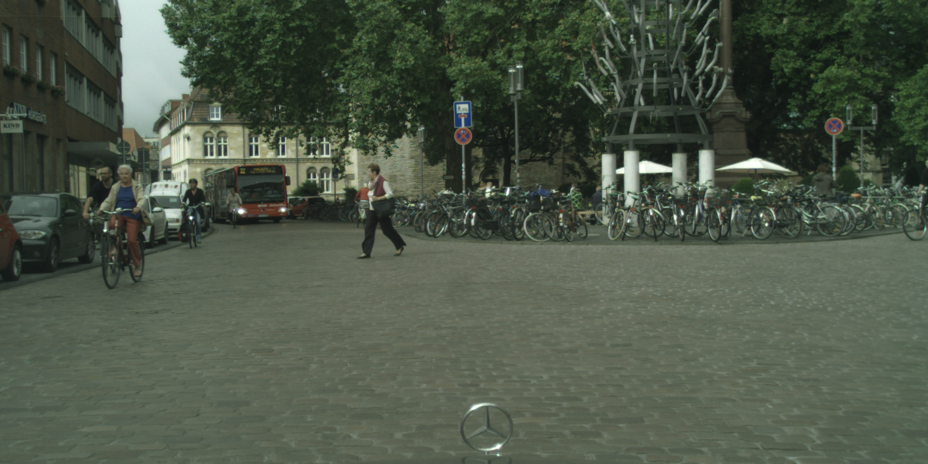}} &
    \subfloat{\includegraphics[width=0.16\textwidth,clip]{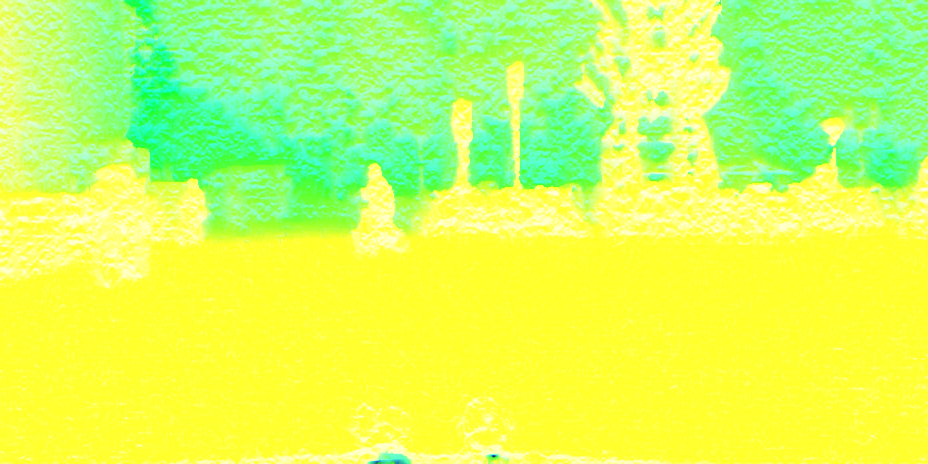}} &
    \subfloat{\includegraphics[width=0.16\textwidth,clip]{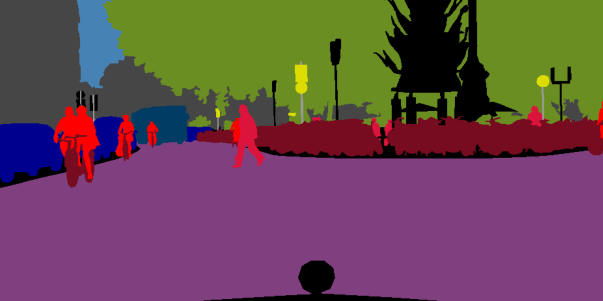}} &
    \subfloat{\includegraphics[width=0.16\textwidth,clip]{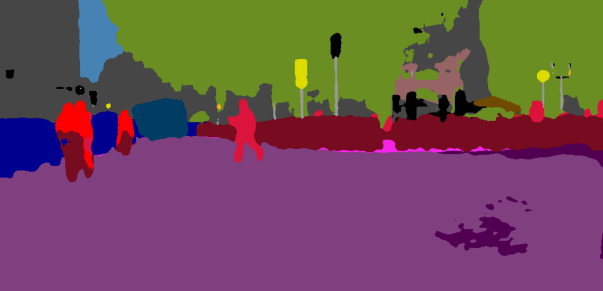}} &
    \subfloat{\includegraphics[width=0.16\textwidth,clip]{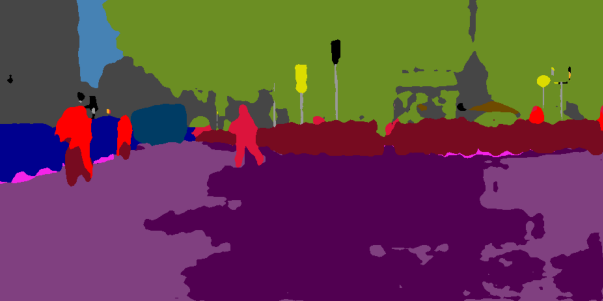}} & 
    \subfloat{\includegraphics[width=0.16\textwidth,clip]{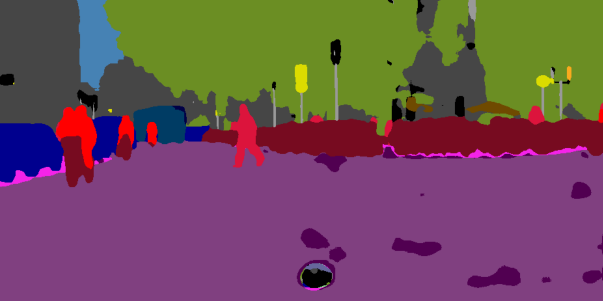}} \vspace*{-1.25em} \\
    \subfloat{\includegraphics[width=0.16\textwidth,clip]{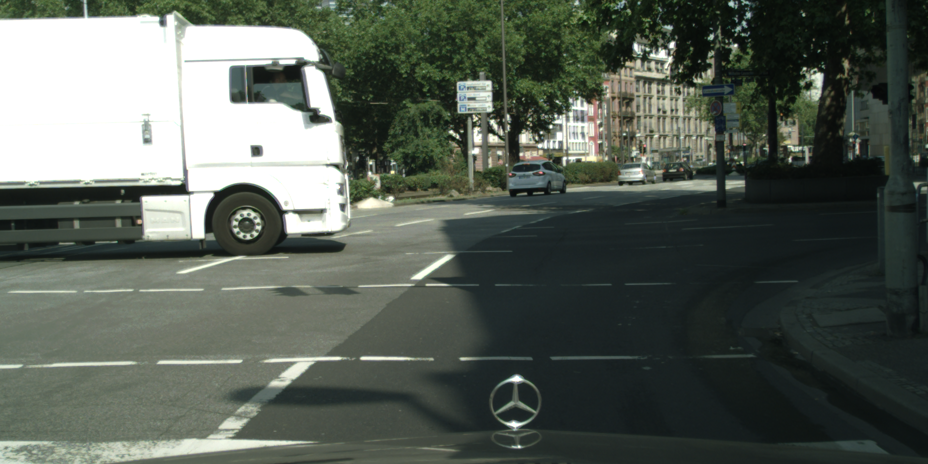}} &
    \subfloat{\includegraphics[width=0.16\textwidth,clip]{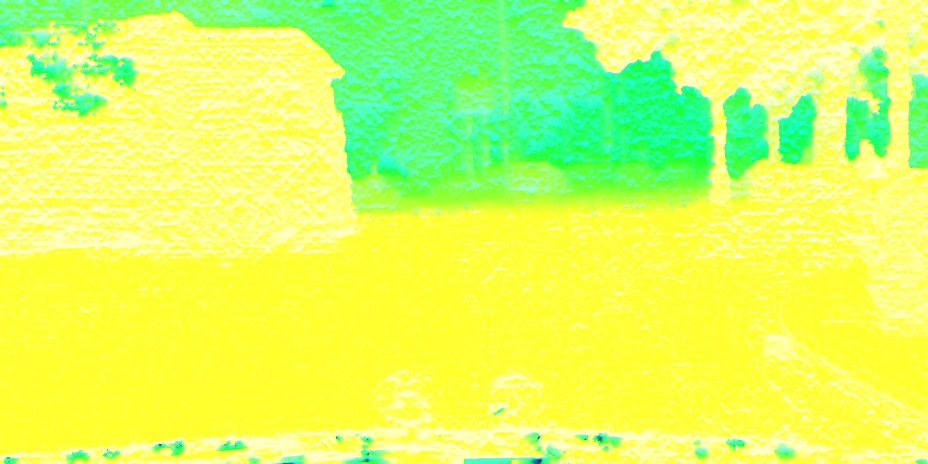}} &
    \subfloat{\includegraphics[width=0.16\textwidth,clip]{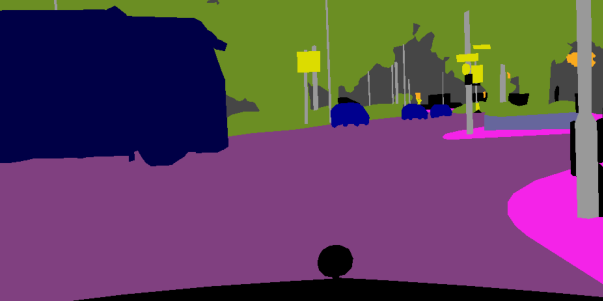}} &
    \subfloat{\includegraphics[width=0.16\textwidth,clip]{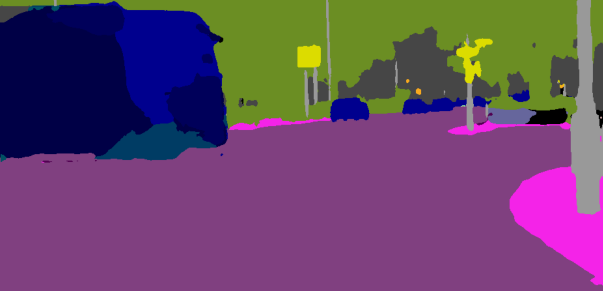}} &
    \subfloat{\includegraphics[width=0.16\textwidth,clip]{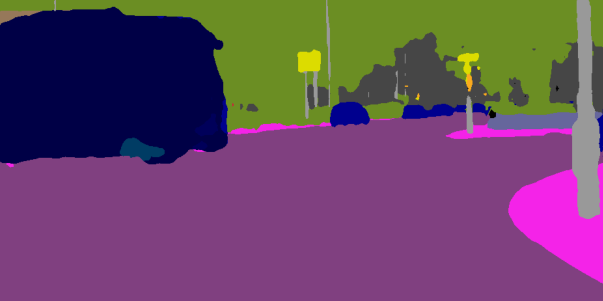}} & 
    \subfloat{\includegraphics[width=0.16\textwidth,clip]{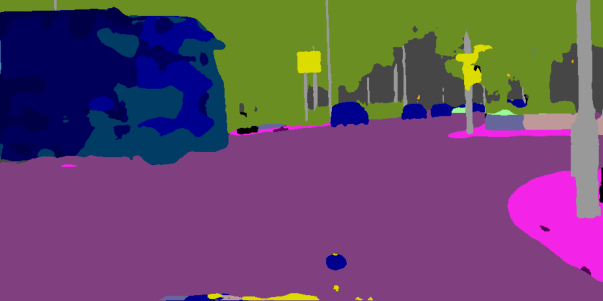}} \vspace*{-0.45em} \\
    \bottomrule
    \end{tabular}
    \caption{\textbf{Qualitative Results of HaarNet.}
    The top three scatter plots show accuracy per image for HaarNet versus SAGate, and select the relative best and worst sample.
    These samples from NYUv2, 2D-3D-S and Cityscapes are shown in the rows with images, interweaving relative best (top, HaarNet outperforms SAGate) and worst (bottom, vice versa) samples.
    In the second row (NYUv2, worst), a reflective surface results in a low-quality depth signal. 
    The morphological HaarNet misclassifies at those depth artefacts, although it does recover the lamp situated on the table, which is not labelled as lamp.
    The third and fourth row show that 2D-3D-S has relatively coarse depth and semantics, complicating learning for all networks. 
    Here, the third best and tenth worst images are shown, since the other failure modes are completely homogeneous images.
    Finally, CityScapes images show that HaarNet is able to classify the road mostly correctly. 
    However, the monochromatic truck poses a challenge.
    Best viewed in colour.
    }
    \label{fig:qualitative-results}
\end{figure}

\begin{table}[t]
    \centering
    \caption{\textbf{Outdoor Segmentation Results.}
    Unlike results from indoor datasets, HaarNet significantly outperforms the baselines.
    It could be this is due to the high amount of structure present in the scenes, which allows absolute operators like morphological ones to make more rigid predictions.
    }
    \resizebox{0.6\textwidth}{!}{
        \begin{tabular}{@{\extracolsep{4pt}}l|ccc@{}}
        \toprule
        Network & mIoU & pixel accuracy & Boundary F1 \\
        \midrule
        Dual ResNet Baseline                            & 0.709 & 0.952 & 0.714 \\
        SAGate \cite{chen2020bi}                        & 0.713 & 0.953 & 0.728 \\
        HaarNet                                         & \bf0.762 & \bf0.961 & \bf0.780 \\
        \bottomrule
        \end{tabular}
    }
    \label{tab:outdoor-results}
\end{table}
\myparagraph{Indoor Semantic Segmentation}
The predictive performance of HaarNet on indoor segmentation datasets is shown in~\autoref{tab:indoor-results}.
For NYUv2, HaarNet outperforms the general baseline and state-of-the-art SAGate on all metrics, provided the networks are trained on the same system, using the same pre-training, and with the same hyper-parameters.
HaarNet is also more efficient, given that it only uses 13.1M additional parameters on top of the encoder, versus 20-25M for the other baselines.
For 2D-3D-S, performance is on par with SAGate.
The main difference between NYUv2 and 2D-3D-S, apart from size, is the quality of the depth signals.
To see so, review the qualitative results, as depicted in~\autoref{fig:qualitative-results}, in which the relative best and worst predictions of HaarNet versus SAGate are shown.
From the centre rows, it can be seen that the depth signals of 2D-3D-S contain many reconstruction artefacts. 
Since morphological operations are more sensitive to geometric modalities, it is possible the depth artefacts complicate learning for HaarNet.

\myparagraph{Outdoor Semantic Segmentation}
HaarNet is also evaluated on the outdoor CityScapes dataset; results are shown in~\autoref{tab:outdoor-results}.
Here, HaarNet significantly outperforms the baselines.
This may be because autonomous driving datasets are much less varied than indoor datasets since traffic is structured to be predictable.
Morphological operators are more calibrated than linear operators, since they act absolutely rather than relatively.
\begin{table}[t]
    \centering
    \caption{\textbf{Ablation Study.}
    Rows show performance due to the morphological up-sampling layers (M-UP), morphological non-linearities (M-ReLU), and the Haar module (MHW).
    Modules are replaced with simple (linear) counterparts when possible.
    Results indicate that, while networks adapt to the available modules, a network that is a full linear-morphological hybrid performs best.
    }
    \resizebox{0.6\textwidth}{!}{
        \begin{tabular}{@{\extracolsep{4pt}}ccc|ccc@{}}
        \toprule
        M-UP & M-ReLU & MHW & mIoU & pixel accuracy & Boundary F1 \\        
        \midrule
        & &                                     & 0.469 & 0.751 & 0.373 \\
        \checkmark & &                          & 0.475 & 0.752 & 0.367 \\
        \checkmark & \checkmark &               & 0.469 & 0.753 & 0.371 \\
        \checkmark & & \checkmark               & 0.481 & 0.755 & 0.363 \\
        \checkmark & \checkmark & \checkmark    & \bf0.510 & \bf0.771 & \bf0.385 \\
        \bottomrule
        \end{tabular}
    }
    \label{tab:ablation-study}
\end{table}

\subsection{Ablation Study} \label{sec:experiments:ablation-study}
The performance due to three core modules of HaarNet are evaluated: morphological up-sampling layers, morphological non-linearities (\hyperref[sec:method:relu-as-dilation]{Section 3.2}), and the morphological Haar Wavelet (MHW) module (\hyperref[sec:method:haar-module]{Section 3.1}).
Results are shown in~\autoref{tab:ablation-study}.
In the ablation study, the base class without any of the introduced modules is architecturally equivalent to a DeeplabV3 \cite{chen2017deeplab}.
All networks exploit ResNet-101 backbones, and are trained for 100 epochs on NYUv2 40-class setting.
Results indicate that the Haar module is important for achieving best performance, since it enables modality fusion at the same time as informed down-sampling.
The ablation also shows that networks are flexible: when modules are replaced by simple (linear) counterparts, the network can come to depend on the reliable DeepLab architecture.

\section{Conclusions} \label{sec:conclusions}
It is the authors' strong belief that the set of operators made available to a network should reflect the admissible transformations on the input modalities.
Morphological operations are more suitable to analysis of geometry-based modalities than convolutions are.
This paper demonstrates that leveraging knowledge from the field of mathematical morphology yields promising results, even when core network design principles are kept similar to those of CNNs.
This was achieved by utilising three core MM concepts and introducing a new pre-training regime, resulting in the large-scale linear-morphological hybrid network called \textbf{HaarNet}.
Further research should address attention mechanisms that enable networks to focus on specific features using modality-specific operators.
This should be explored both across scales and along feature depth; the latter has yet to be investigated in the literature.

{\small
\bibliographystyle{splncs04}
\bibliography{refs}

\begin{thebibliography}{10}
\providecommand{\url}[1]{\texttt{#1}}
\providecommand{\urlprefix}{URL }
\providecommand{\doi}[1]{https://doi.org/#1}

\bibitem{armeni2017joint}
Armeni, I., Sax, S., Zamir, A.R., Savarese, S.: Joint 2d-3d-semantic data for indoor scene understanding. arXiv preprint arXiv:1702.01105  (2017)

\bibitem{burgeth2005logarithmic}
Burgeth, B., Weickert, J.: An explanation for the logarithmic connection between linear and morphological system theory. IJCV  \textbf{64},  157--169 (09 2005)

\bibitem{charisopoulos2017morphological}
Charisopoulos, V., Maragos, P.: Morphological perceptrons: geometry and training algorithms. In: ISMM. pp. 3--15. Springer (2017)

\bibitem{chen2013deep}
Chen, B., Polatkan, G., Sapiro, G., Blei, D., Dunson, D., Carin, L.: Deep learning with hierarchical convolutional factor analysis. TPAMI  \textbf{35}(8),  1887--1901 (2013)

\bibitem{chen2017deeplab}
Chen, L.C., Papandreou, G., Kokkinos, I., Murphy, K., Yuille, A.L.: Deeplab: Semantic image segmentation with deep convolutional nets, atrous convolution, and fully connected {CRF}s. TPAMI  \textbf{40}(4),  834--848 (2017)

\bibitem{chen2021global}
Chen, S., Zhu, X., Liu, W., He, X., Liu, J.: Global-local propagation network for rgb-d semantic segmentation. arXiv preprint arXiv:2101.10801  (2021)

\bibitem{chen2020bi}
Chen, X., Lin, K.Y., Wang, J., Wu, W., Qian, C., Li, H., Zeng, G.: Bi-directional cross-modality feature propagation with separation-and-aggregation gate for rgb-d semantic segmentation. In: ECCV. pp. 561--577. Springer (2020)

\bibitem{cordts2016cityscapes}
Cordts, M., Omran, M., Ramos, S., Rehfeld, T., Enzweiler, M., Benenson, R., Franke, U., Roth, S., Schiele, B.: The cityscapes dataset for semantic urban scene understanding. In: CVPR (2016)

\bibitem{csurka2004good}
Csurka, G., Larlus, D., Perronnin, F., Meylan, F.: What is a good evaluation measure for semantic segmentation? TPAMI  \textbf{26}(1) (2004)

\bibitem{dimitriadis2020advances}
Dimitriadis, N., Maragos, P.: Advances in the training, pruning and enforcement of shape constraints of morphological neural networks using tropical algebra. arXiv preprint arXiv:2011.07643  (2020)

\bibitem{franchi2020deep}
Franchi, G., Fehri, A., Yao, A.: Deep morphological networks. Pattern Recognition  \textbf{102},  107246 (2020)

\bibitem{fukushima1975cognitron}
Fukushima, K.: Cognitron: A self-organizing multilayered neural network. Biological cybernetics  \textbf{20}(3-4),  121--136 (1975)

\bibitem{groenendijk2022morphpool}
Groenendijk, R., Dorst, L., Gevers, T.: Morphpool: Efficient non-linear pooling \& unpooling in cnns. In: 33rd British Machine Vision Conference 2022, {BMVC} 2022, London, UK, November 21-24, 2022. {BMVA} Press (2022)

\bibitem{groenendijk2022backprop}
Groenendijk, R., Dorst, L., Gevers, T.: Geometric back-propagation in morphological neural networks. TPAMI pp.~1--8 (2023)

\bibitem{gupta2014learning}
Gupta, S., Girshick, R., Arbel{\'a}ez, P., Malik, J.: Learning rich features from rgb-d images for object detection and segmentation. In: ECCV. pp. 345--360. Springer (2014)

\bibitem{he2016deep}
He, K., Zhang, X., Ren, S., Sun, J.: Deep residual learning for image recognition. In: CVPR. pp. 770--778 (2016)

\bibitem{heijmans2000nonlinear}
Heijmans, H.J., Goutsias, J.: Nonlinear multiresolution signal decomposition schemes. ii. morphological wavelets. TIP  \textbf{9}(11),  1897--1913 (2000)

\bibitem{heijmans1991morphological}
Heijmans, H.J., Toet, A.: Morphological sampling. CVGIP: Image understanding  \textbf{54}(3),  384--400 (1991)

\bibitem{hernandez2020hybrid}
Hern{\'a}ndez, G., Zamora, E., Sossa, H., T{\'e}llez, G., Furl{\'a}n, F.: Hybrid neural networks for big data classification. Neurocomputing  \textbf{390},  327--340 (2020)

\bibitem{hu2018squeeze}
Hu, J., Shen, L., Sun, G.: Squeeze-and-excitation networks. In: CVPR. pp. 7132--7141 (2018)

\bibitem{jarrett2009best}
Jarrett, K., Kavukcuoglu, K., Ranzato, M., LeCun, Y.: What is the best multi-stage architecture for object recognition? In: ICCV. pp. 2146--2153. IEEE (2009)

\bibitem{jiang2022multi}
Jiang, S., Xu, Y., Li, D., Fan, R.: Multi-scale fusion for rgb-d indoor semantic segmentation. Scientific Reports  \textbf{12}(1),  20305 (2022)

\bibitem{mondal2022morphological}
Mondal, R., Santra, S., Mukherjee, S.S., Chanda, B.: Morphological network: How far can we go with morphological neurons? In: BMVC. {BMVA} Press (2022)

\bibitem{paszke2019pytorch}
Paszke, A., Gross, S., Massa, F., Lerer, A., Bradbury, J., Chanan, G., Killeen, T., Lin, Z., Gimelshein, N., Antiga, L., Desmaison, A., Kopf, A., Yang, E., DeVito, Z., Raison, M., Tejani, A., Chilamkurthy, S., Steiner, B., Fang, L., Bai, J., Chintala, S.: Pytorch: An imperative style, high-performance deep learning library. In: NeuRIPS, pp. 8024--8035. Curran Associates, Inc. (2019)

\bibitem{qiu2018frelu}
Qiu, S., Xu, X., Cai, B.: Frelu: flexible rectified linear units for improving convolutional neural networks. In: ICPR. pp. 1223--1228. IEEE (2018)

\bibitem{ritter1996introduction}
Ritter, G.X., Sussner, P.: An introduction to morphological neural networks. In: ICPR. vol.~4, pp. 709--717. IEEE (1996)

\bibitem{roy2021morphological}
Roy, S.K., Mondal, R., Paoletti, M.E., Haut, J.M., Plaza, A.: Morphological convolutional neural networks for hyperspectral image classification. IEEE Journal of Selected Topics in Applied Earth Observations and Remote Sensing  \textbf{14},  8689--8702 (2021)

\bibitem{serra1982image}
Serra, J.: Image analysis and mathematical morphology  (1983)

\bibitem{serra1992overview}
Serra, J., Vincent, L.: An overview of morphological filtering. Circuits, Systems and Signal Processing  \textbf{11}(1),  47--108 (1992)

\bibitem{shen2019deep}
Shen, Y., Shih, F.Y., Zhong, X., Chang, I.C.: Deep morphological neural networks. PRAI  \textbf{36}(12),  2252023 (2022)

\bibitem{silberman2012indoor}
Silberman, N., Hoiem, D., Kohli, P., Fergus, R.: Indoor segmentation and support inference from {RGBD} images. ECCV  \textbf{7576},  746--760 (2012)

\bibitem{sussner1998morphological}
Sussner, P.: Morphological perceptron learning. In: Proceedings of the 1998 ISIC/CIRA. pp. 477--482. IEEE (1998)

\bibitem{sutskever2013importance}
Sutskever, I., Martens, J., Dahl, G., Hinton, G.: On the importance of initialization and momentum in deep learning. In: ICML. pp. 1139--1147. PMLR (2013)

\bibitem{velasco2022morphoactivation}
Velasco-Forero, S., Angulo, J.: Morphoactivation: Generalizing relu activation function by mathematical morphology. In: DGMM. pp. 449--461. Springer (2022)

\bibitem{velasco-forero2022fixed}
Velasco-Forero, S., Rhim, A., Angulo, J.: Fixed point layers for geodesic morphological operations. In: BMVC. {BMVA} Press (2022)

\bibitem{xie2014spatial}
Xie, L., Tian, Q., Wang, M., Zhang, B.: Spatial pooling of heterogeneous features for image classification. TIP  \textbf{23}(5),  1994--2008 (2014)

\bibitem{yosinski2014transferable}
Yosinski, J., Clune, J., Bengio, Y., Lipson, H.: How transferable are features in deep neural networks? NeuRIPS  \textbf{27} (2014)

\bibitem{zeiler2014visualizing}
Zeiler, M.D., Fergus, R.: Visualizing and understanding convolutional networks. In: ECCV. pp. 818--833. Springer (2014)

\bibitem{zeiler2011adaptive}
Zeiler, M.D., Taylor, G.W., Fergus, R.: Adaptive deconvolutional networks for mid and high level feature learning. In: ICCV. pp. 2018--2025. IEEE (2011)

\bibitem{zhou2022canet}
Zhou, H., Qi, L., Huang, H., Yang, X., Wan, Z., Wen, X.: Canet: Co-attention network for rgb-d semantic segmentation. Pattern Recognition  \textbf{124},  108468 (2022)

\end{thebibliography}
}

\end{document}